\documentclass[review]{elsarticle}

\usepackage{amsmath,amssymb,amsfonts}
\usepackage{algorithmic}
\usepackage{graphicx}
\usepackage{textcomp}
\usepackage{url,microtype,subcaption}
\usepackage[onehalfspacing]{setspace}
\usepackage{graphics} 
\usepackage{epsfig} 
\usepackage{mathptmx} 
\usepackage{times} 
\usepackage{lipsum}
\usepackage{multicol}
\usepackage{pdfpages}
\usepackage{epstopdf}
\usepackage{multirow}
\usepackage{booktabs}
\usepackage{afterpage}
\usepackage{soul}
\usepackage{graphicx}
\usepackage{hyperref}
\usepackage{color,soul}
\usepackage{xcolor}
\hypersetup{
     colorlinks = true,
     linkcolor = blue,
     anchorcolor = blue,
     citecolor = green,
     filecolor = magenta,
     urlcolor = cyan
     }

\journal{Journal of \LaTeX\ Templates}

\begin{document}

\begin{frontmatter}

\title{Exploring Thermal Images for Object Detection in Underexposure Regions for Autonomous Driving}

\author[mainadd]{Farzeen Munir }
\author[mainadd]{Shoaib Azam }
\author[mainadd]{Muhammd Aasim Rafique }
\author[secadd]{Ahmad Muqeem Sheri }
\author[mainadd]{Moongu Jeon \corref{mycorrespondingauthor}}
\cortext[mycorrespondingauthor]{Corresponding author}
\ead{mgjeon@gist.ac.kr}
\author[thirdadd]{Witold Pedrycz}



\address[mainadd]{School of Electrical Engineering and Computer Science, Gwangju Institute of Science and Technology, Gwangju, South Korea}
\address[secadd]{Department of Computer Software Engineering, National University of Sciences and Technology (NUST), Islamabad, Pakistan} 
\address[thirdadd]{Department of Electrical and Computer Engineering, University of Alberta, Edmonton, AB T6R 2V4, Canada, with the Department of Electrical and Computer Engineering, Faculty of Engineering, King Abdulaziz University, Jeddah 21589, Saudi Arabia, and also with the Systems Research Institute, Polish Academy of Sciences, Warsaw 01-447, Poland.}






\begin{abstract}
Underexposure regions are vital to construct a complete perception of the surroundings for safe autonomous driving. The availability of thermal cameras has provided an essential alternate to explore regions where other optical sensors lack in capturing interpretable signals. A thermal camera captures an image using the heat difference emitted by objects in the infrared spectrum, and object detection in thermal images becomes effective for autonomous driving in challenging conditions. Although object detection in the visible spectrum domain imaging has matured, thermal object detection lacks effectiveness. A significant challenge is scarcity of labeled data for the thermal domain which is desiderata for SOTA artificial intelligence techniques. This work proposes a domain adaptation framework which employs a style transfer technique for 
transfer learning from visible spectrum images to thermal images. The framework uses a generative adversarial network (GAN) to transfer the low-level features from the visible spectrum domain to the thermal domain through style consistency. The efficacy of the proposed method of object detection in thermal images is evident from the improved results when used styled images from  publicly available thermal image datasets (FLIR ADAS and KAIST Multi-Spectral).

\end{abstract}


\end{frontmatter}


\section{Introduction}
Object detection, as one of the elemental component of the perception system, has a wide range of application ranging from medical to autonomous driving. For autonomous driving, the perception of the environment plays a pivotal role in determining the safety of the autonomous driving. Environmental perception is generally defined as awareness of or knowledge about the surroundings and the understanding of the situation by the visual perception \cite{pr}. Since the autonomous driving has to offer broader access to mobility, the safety standards as instructed by SOTIF (Safety of the intended functionality)\footnote{https://newsroom.intel.com/wp-content/uploads/sites/11/2019/07/Intel-Safety-First-for-Automated-Driving.pdf} perception system constitute of the object detection must reflect the safe and secure course of action for the autonomous driving.


\par
The sensors commonly used for perception in the autonomous driving includes Lidar, RGB cameras, and radar. Object detection using these sensor modalities provides the perception for the autonomous driving, but in contrast, each of these sensor modalities has its drawbacks.
Lidar gives a sparse 3D map representation of the environment, but small objects like pedestrians and cyclists are hard to detect at a large distance. Similarly, the RGB camera performs poorly in unfavorable illumination conditions such as low lighting, sun glare, and glare from the headlight of the vehicle. Radar has a low spatial resolution to detect pedestrians accurately. There exists a performance gap in object detection for adverse lighting conditions \cite{kt}. The inclusion of a thermal camera in the sensor's suite provides a way to fill the blind spots in environmental perception.  The thermal camera is robust against illumination variation and has advantage to be deployed during day and night. The object detection and classification are indispensable for visual perception, which provides a basis for computing perception in an autonomous driving.
\begin{figure*}[t]
      \centering
      \includegraphics[width=12cm]{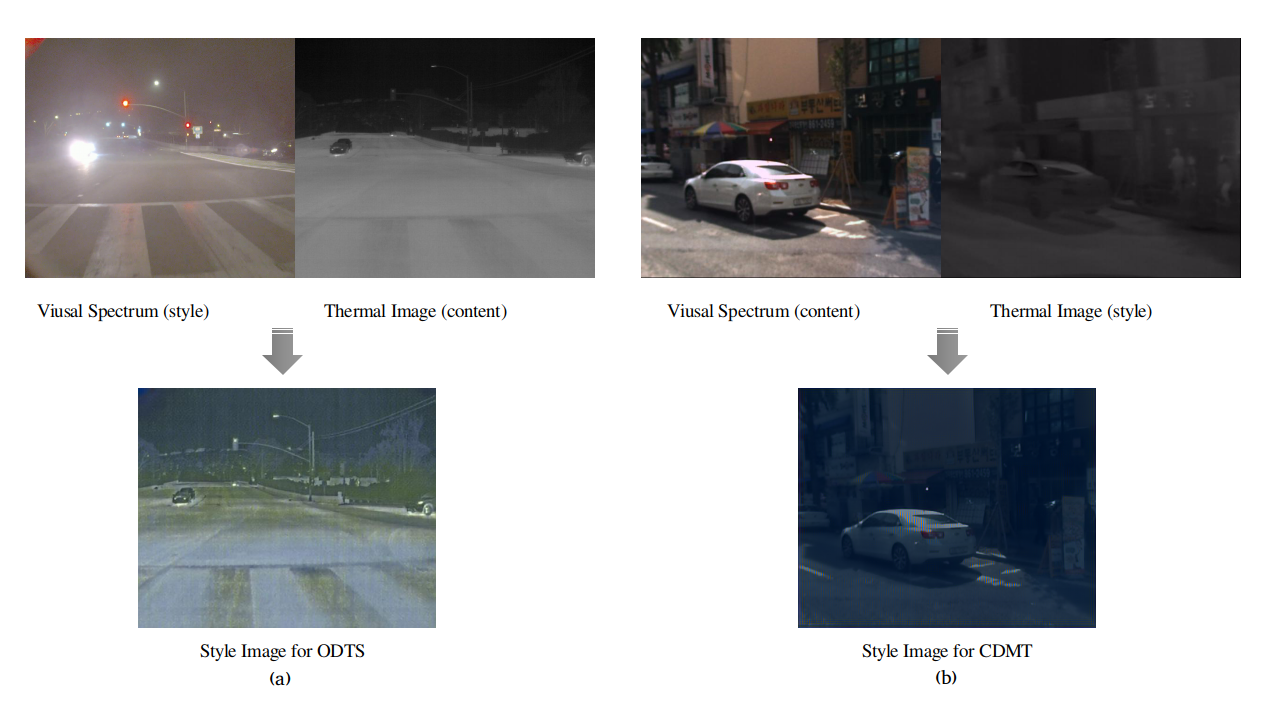}
      \caption{(a) Object detection in thermal images through style consistency (ODSC). Visible spectrum (RGB image) is treated as a style image whereas, the thermal image is considered as content image. The output shows the enhanced image having low-level features adapt from the visible spectrum. (b) Cross-domain model transfer with style transfer. Style from the thermal image is transferred to the visible spectrum (RGB content image).}
      \label{style-images}
\end{figure*}
\par
Object detection in visible spectrum (RGB) domain is considered sufficient for conventional AI applications, and has resulted in deep neural network models for robust object detection \cite{faster-rcnn} \cite{ssd} \cite{yolo}. However, the accuracy of object detection in thermal images has not yet attained the state-of-the-art results compared to its counterpart visible spectrum RGB images.
The aforementioned object detection algorithms depend on networks that have been trained on sizable RGB datasets such as ImageNet \cite{image}, PASCAL-VOC \cite{pascal}, and MS-COCO \cite{coco}. There exist a comparable scarcity of such large scale public datasets in the thermal domain. Two primary datasets for urban thermal imagery that are publicly available include, FLIR-ADAS image dataset\footnote{https://www.flir.in/oem/adas/adas-dataset-form/} and KAIST Multi-Spectral dataset \cite{KAIST}. KAIST Multi-Spectral dataset only gives annotations for persons, while the FLIR-ADAS dataset gives annotation for four classes. In order to overcome the absence of the large scale labeled dataset, here, a domain adoption framework for object detection in the thermal domain is presented.
\par
Currently, numerous approaches for domain adaptation have been introduced, which aims to narrow down the gap between source and target domain. Among many, generative adversarial networks (GAN) \cite{gan} and domain adaptation \cite{domain} for the feature adaptation are noteworthy. The domain adaptation prospects in data starved thermal images domain which is motivation of this study. It explores a derivative of closing the gap between visible and infrared spectrum in the context of object detection. Domain adaptation is influenced by generative models, for instance, CycleGAN \cite{cyclegan} that translates the single instance of source domain to target domain without translating the style attributes to the target domain. The low-level visual cues have an implicit impact on the performance of object detection \cite{low}. The delegation of these visual cues in the target domain from the source domain can be beneficial for robust object detection in the target domain.
\par 
In this work, we have proposed a framework based on domain adaptation for thermal object detection by translating the low-level features adopted from a source domain (RGB) to a target domain (thermal).
A multi-style transfer approach is employed in the domain adaptive framework for the translation of low-level features such as curvatures and edges from the source domain to the target domain.
Deep learning-based object detection architectures that rely on classical backbone like VGG \cite{vgg}, ResNet \cite{resnet}, are trained on the multi-style transfer images from scratch for the robust object detection in the thermal domain (target domain).
Moreover, we have proposed a cross-domain model transfer \footnote{ The cross-domain model is coined by the cross-domain interoperability where the systems from different domain interacts in information exchange, service or work together to achieve the common goal. The cross domain model is the knowledge transfer of model that is trained in one domain and can be used in other domain by implying the feature learned by that model is reused for the other domain.} method for object detection in thermal images supplementing the domain adaptation. The cross-domain model transfer for which the object detection deep neural networks have trained in the source domain (visible spectrum). The trained models, referred to as cross-domain models, are evaluated with multi-style transfer images and without multi-style transfer images in the target domain (infrared spectrum). The proposed techniques are evaluated on FLIR-ADAS and KAIST Multi-Spectral \cite{KAIST} datasets, and PASCAL-VOC evaluation is used to determine the average mean precision of the detected objects\cite{pascal}. 
The major contributions in this work are highlighted below:
\begin{enumerate}
\item Fusion of two domains at the data level for the object detection and confirming the hypothesis by the extensive experimentation using the available FLIR ADAS and KAIST Multi-Spectral datasets. The underlying thesis is that the style transfer relegate low frequency features from source domain to target domain that form  the basis of improved accuracy of detection and classification. 
\item Improved object detection in the infrared spectrum (thermal images) by exploring the low-level features through style consistency. The proposed object detection framework outperformed existing benchmarks in terms of mean average precision. 
\item Cross-domain model transfer paradigm not only enhances the object detection in the infrared spectrum (thermal images) but also provides an alternative yet effective method for labeling the unlabeled dataset. 
\end{enumerate}
\par 
This work illustrates a novel approach to improve object detection for thermal images is introduced by transferring knowledge through domain adaptation employing style transfer. This work's main motivation is to handle the scarcity or non-existence of labeled data, which is an utmost challenge to the research community, and further, the labeling of data is an expensive task.
\par
The paper is organized as follows: Section II discusses the related literature. In Section III, the proposed methodology is discussed. Section IV focuses on experimentation and analysis of results. Section V shows the comparison and discussion about the proposed method. Section VI concludes the study.

\section{Related Work}
\subsection{Object Detection}
 Human vision is capable to identify objects in countless challenging conditions, but it is not a trivial task for the autonomous driving.  The ultimate goal of object detection in images is to localize and identify all instances of the same object or different objects present in the image. 
 Significant work is done on person detection in thermal images by considering the temperature difference between the hot body and cool surrounding. Classical image processing techniques can be used for detection, like thresholding is used in \cite{a1}. They have formulated the threshold value based on a model, which considers different thermal images' characteristics. 
 The Histogram of oriented gradient (HOG) features and local binary patterns (LBP) are used to extract features from thermal images, and the features are used to train the Support Vector Machine (SVM) classifiers in \mbox{\cite{a2}}. \mbox{\cite{a3}} used HOG features combined with geometric features such as mean and contrast to compute a set of features that are then used to train the SVM classifier. The classical methods lack robust features and accuracy in detecting thermal object detection as compared to deep neural networks and are not suitable for dynamic situation of autonomous drivings.  Deep neural networks have gained reputation in object detection tasks for RGB images and are used for object detection in thermal images\cite{q4}. In \mbox{\cite{a5}}, they first train two separate convolution network on thermal and RGB images separately. Then they proposed four fusion architecture which integrates two convolution network at different stages of convolution. They discover that convolution neural network train on thermal images and RGB images provide complementary information on discriminating objects in thermal images and thus yield better performance. Similar work is conducted in \mbox{\cite{a6}} where they have proposed fusion architecture to study the benefit of using multispectral data for thermal object detection. \mbox{\cite{a7}} have proposed a real-time multispectral pedestrian detector by training You Only Look Once (YOLO) object detector with the input of 3 RGB channels in addition to thermal as to the fourth channel. \cite{q5} proposed a method based on fusion of thermal and visible domain using target enchanced multi-scale decomposition model. The Laplacian pyramid is used to compute low-frequency feature in thermal images  and than fuse the information with visible spectrum to improve the features of target object, which improve the reliability of target recognition and detection.  

 \subsection{Domain Adaptation}
Typically, neural networks encounter performance degradation when they are tested upon different datasets due to environmental changes. In some cases, the dataset is not large enough to train and optimize a network. Therefore techniques like domain adaptation provide a crucial tool to the research community\cite{q1}. 
\par
The domain adaptation for object detection includes techniques like the generation of synthetic data or augmentation to real data to train the network. \cite{a11} have used publicly available object detection labeled datasets coming from various domains and multiple classes and merged them. For example, the fashion dataset Modanet is merged with the MS-COCO dataset by leveraging Faster-RCNN using domain adaptation.  In \cite{a12}, Faster-RCNN is used to make image and instance-level adaptation. \cite{a13} have introduced a two-step method, where they have optimized a detector to low-level features, and then it is developed as a robust classifier for high-level features by enforcing distance minimization between content and style image. \cite{a14} has proposed a cross-domain semi-supervised learning structure that takes advantage of pseudo annotations to learn optimal representations of the target domain. They have used the fine-grained domain transfer, progressive confidence based annotation augmentation, and annotation sampling strategy.

\subsection{Transfer Learning}
In real-world applications, the training and test data do not belong to the same feature-space or have similar data distributions, although most machine learning algorithms hold this assumption . In light of violation of this assumption, most machine learning models need to be rebuilt using new labeled training data \cite{q2}. For such task transfer learning helps transfer the knowledge between task domains\cite{ma}. \cite{mb} has exhibited the transfer learning-based framework for object detection datasets with a very few training examples. They have augmented the examples from each class by importing the examples from other classes and transforming them to be more similar to the target class. \cite{mc} presents a boosting framework to transfer learning from multiple sources. The brute force transfer of knowledge might transfer weak relationships, which reduces the performance of the classifier. The knowledge is borrowed from multiple sources to evade negative transfer. \cite{md} performs a study to examine the efficacy of transfer learning affected by the choice of dataset. They have proposed adaptive transfer learning, a simple and effective pre-training technique based on weights computed on the target dataset. \cite{me} solves the fine-grained visual categorization problem using domain adaptive transfer learning. They have fed the neural network additional data by augmenting the data through a visual attention mechanism and then fine-tune it on the base network. \cite{q3}  propose a new technique based on transfer learning to relegate the knowledge from source task to the target task containg uncertain labels. 

 \subsection{Style Transfer}
Image Style transfer is a process that renders the content of the image from one domain with the style of another image from another domain. \cite{a15} has demonstrated the use of feature representation from the convolution neural network for style transfer between two images. They have shown that features obtained from CNN are separable. They manipulate the feature representation between style and content images to generate new and visually meaningful images. \cite{a16} have proposed style transfer based on a single object. They have used patch permutation to train a GAN to learn the style and apply it to the content image. \cite{a17} has introduced XGAN, consisting of auto-encoder, which captures the shared features from style and content images in an unsupervised way and along which it learns the translation of style onto the content image. \cite{a18} has proposed the CoMatch layer, which learns the second-order statistics of features and then matches them with the style image. Using the CoMatch layer, they have developed the  Multi-style Generative Network giving a real-time performance.
\par
There is still a need for improvement in thermal object detection in the context of the aforementioned related literature extending from object detection, transfer learning, style transfer, and domain adaptation. The resurgence of feature extraction without human supervision has greatly improved by the deep neural networks in the visible spectrum RGB domain for the classification, detection and prediction problems. In addition, the leverage of the proposed approach is to perform domain adaptation for other datasets, like introducing foggy weather in the KITTI dataset \cite{kitti} or convert day images to night images. 

\section{Proposed Method}
This section presents the proposed domain adaptive framework for thermal object detection from visible RGB domain to thermal domain.

\subsection{Object Detection in Thermal Images through Style Consistency (ODSC)}
The recent advances in deep learning have revolutionized object detection in the visible RGB image domain. However, in the thermal image domain, there is still room for improvement.  Deep neural networks as function approximators perform low-level and high-level feature extraction for the possible classification/prediction problem \mbox{\cite{deg} \cite{low}}. 
Here, we argue that transferring the low-level features from the source domain (RGB) using domain adaption increases the target domain's (thermal) object detection performance.
\par
\begin{figure*}[t]
      \centering
      \includegraphics[width=13cm]{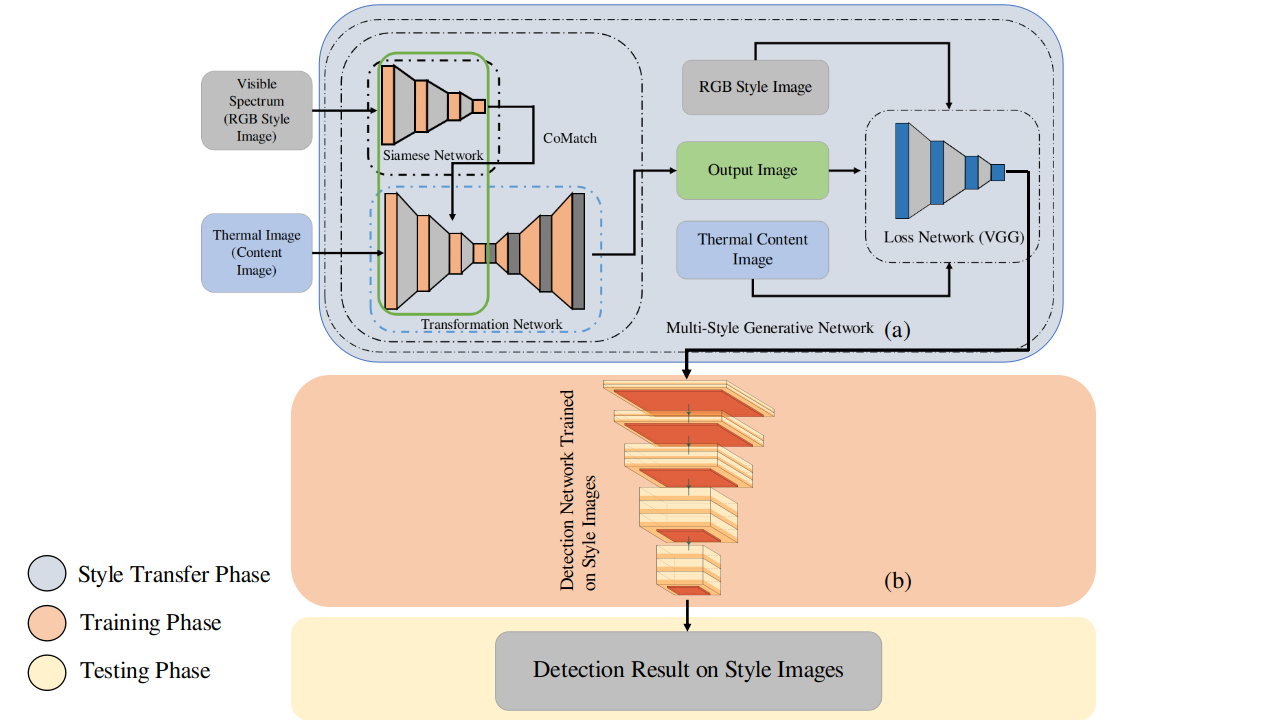}
      \caption{The proposed model framework for object detection in thermal images through style consistency. (a) Multi-style generative network architecture for generating the style images. Visible spectrum (RGB images) and thermal images are given as style and content image respectively to the network. The siamese network captures the low-level features of style image, which is transferred to the transformation network through the CoMatch layer. A pre-trained loss network is used for MSGNet learning by computing the difference between content and style image with the targets. (b) The detection networks which includes (Faster-RCNN backbone with ResNet-101, SSD-300 with backbone VGG16, MobileNet, and EfficientNet, SSD-512 with VGG16 backbone) are trained on the style images and then tested in the target domain (thermal images) for the object detection.}
      \label{proposed}
\end{figure*}
The knowledge transfer using the domain adaptation between the thermal image (content images $x_c$) and visible spectrum (RGB) images (style images $x_s$), we have adopted the multi-style generative network (MSGNet) for style transfer \mbox{\cite{a18}}. The leverage of translating the specific style from the source to the target domain through the multi-style generative network provides an extra edge over the CycleGAN \cite{cyclegan}. The CycleGAN generates one translated image from the source image of a specific style. MSGNet provides the capability to translate multi-style from the source domain to the target domain while closing the gap between the two domains. The network extracts low-level features such as texture and edges from the source domain while keeping the high-level features consistent in the target domain. Fig. \ref{proposed}(a) shows the framework for transferring the style from the visible spectrum (RGB) images to thermal images.
\par 
The architecture of the MSGNet is shown in Fig. {\ref{proposed}}(a). MSGNet network takes both the content image $x_c$ and style image $x_s$ as input, while the previously known architectures, like, Neural Style \mbox{\cite{a16}} that takes only the content image and then generates the transferred image. The Generator network $(G)$ is composed of an encoder consisting of the siamese network \mbox{\cite{sia}}, which shares its network weights with the transformation network through the CoMatch layer. The CoMatch layer matches the second-order feature statistics of content image $x_c$ to the style images $x_s$. For a given content image and a style image, the activation of the descriptive network at the $j^{th}$ scale $\mathcal{F}^j(x) \in \mathbb{R}^{C_j\times H_j\times W_j}$ represents the content image $x_c$ where $C_j$, $H_j$, $W_j$ are the number of feature map channels, the height of feature map and width respectively.
The distribution of features in style image $x_s$ is represented using the Gram Matrix $\mathcal{G}(\mathcal{F}^j(x)) \in \mathbb{R}^{C_j\times C_j}$ given by Eq. {\ref{equ1}}.  In order to find the desired solution in the CoMatch layer that preserves the semantic content of source image as well as matches the feature statics of target style, an iterative approximation approach is adopted by incorporating the computational cost in the training stage as shown in the Eq. {\ref{equ2}}. 
\par 
The minimization of a weighted combination of the content and style difference between the generator network output and targets for a given pre-trained loss network $\mathcal{F}$. The generator network is given by $G(x_c,x_s)$ and parameterized by $W_G$, (weights). The learning is done by sampling the content image $x_c \sim X_c$ and style image $x_s \sim X_s$, and estimating the weights, $W_G$ of the generator $G(x_c,x_s)$ to minimize the loss: 

\begin{equation}
\hspace{-4cm}
 \label{equ3}
    \begin{aligned}
    & A = \lambda _{c} \left \| \mathcal{F}_{x_c}(G(x_c,x_s))-\mathcal{F}_{x_c}(x_c) \right \|^2_F\; ,&& \\
    & B = \lambda _{s}\sum_{j=1}^{K}\left \| \mathcal{G}(\mathcal{F}^j((G(x_c,x_s)))- \mathcal{G}(\mathcal{F}^j(x_s)) \right \|^2_F \; , && \\
    & C = \lambda_{TV}l_{TV}(G(x_c,x_s))\; ,&& \\
     & \hat{W}_{G}=argmin E_{x_c,x_s}\left \{ A + B + C \right \} \; ,&&
    \end{aligned}
\end{equation}

\noindent where $\lambda _{c}$ and $\lambda _{s}$ are the regularization parameters for content and style losses. The content image is consided at scale $c$ and style image is considered at scales $i \in {1,...,K}$. The total variational regularization is $l_{TV}$, which is used for the smoothness of the generated image \mbox{\cite{ph}}. 
\par
\begin{align}
\label{equ1}
\hspace{-2.7cm}
    \mathcal{G}(\mathcal{F}^j(x))=\Phi(\mathcal{F}^j(x))\Phi(\mathcal{F}^j(x))^T \; ,&&
\end{align}
where $\Phi$ is a reshaping function in Gram Matrix $\mathcal{G}$ for zero-centered data. 

\begin{align}
\label{equ2}
\hspace{-2cm}
\hat{y}^j = \Phi^{-1}\left [ \Phi(\mathcal{F}^j(x_c)^T)W\mathcal{G}(\mathcal{F}^j(x_s)))\right ]^T \; ,&&
\end{align}
where $W$ is a learnable matrix.
\par
The proposed framework for object detection through style consistency is presented in Fig. {\ref{proposed}}. It illustrates that the network consists of two modules; the first part consists of a multi-style network. It generates the style images by adapting low-level features transformation between the content image consisting of thermal image and style image consisting of the RGB image.  As compared to the thermal images, the transferred style images contain low-level features, but the semantic shapes are preserved in these generated images keeping the high-level semantic features consistent. The second module is comprised of the state-of-the-art detection architectures: Faster-RCNN \mbox{\cite{faster-rcnn}} backbone with ResNet-101 \mbox{\cite{resnet}}, SSD-300 and 512 \mbox{\cite{ssd}} with backbone VGG16 \mbox{\cite{vgg}}, MobileNet \mbox{\cite{mobilenet}} and EfficientNet \mbox{\cite{enet}}.  The networks are trained on the styled images, which bridge the gap between the visible spectrum RGB images and thermal images.  
The trained detection network is evaluated on  thermal images. The accuracy of testing on thermal images shows the efficacy of object detection. 

\section{Experimentation and Results}

\subsection{Datasets}
In this study, we have used two thermal image datasets. First is the FLIR-ADAS dataset, and the second one is the KAIST Multi-Spectral dataset \cite{KAIST}. FLIR-ADAS dataset consists of $10228$ images with objects annotated using a bounding box as an evaluation measure. The objects are classified into four categories, i.e., car, person, bicycle, and dog. However, the dog category has very few annotations, so it is not considered in this study. The images have a resolution of $640\times 512$ and obtained from FLIR Tau2 Camera. The dataset consists of day and night images, approximately $60\% $ $(6136) $ images are captured during the daytime, and $40\% $ $(4092)$ images are captured during nighttime. The dataset consists of both visible spectrum (RGB images) and thermal images, but annotations are only available for thermal images. The visible spectrum (RGB images) and thermal images are not paired so that the thermal annotations cannot be used with a visible spectrum (RGB images). Thermal images with annotations are only considered in this study. A standard split \footnote{As given by FLIR ADAS repository } of the dataset into training and validation data is considered during experimentation. The training dataset consists of  $8862$ images, and the validation contains $1366$ images, as shown in Table-{\ref{table23}}.
\par
The KAIST Multi-Spectral dataset contains $95000$ images from both the visible spectrum (RGB images) and the thermal spectrum, and for each category, the dataset has both daytime and nighttime images. Annotations are only provided for the person class with a given bounding box.
The visible spectrum (RGB images) and thermal images are paired, which means annotations for the thermal and the visible spectrum (RGB images) are the same. Images are captured using a FLIR A35 camera with a resolution of $320 \times 256$. We have applied a standard split \footnote{As given by KIAST repository.} of the dataset, using $76000$ of the images in the dataset in training and  $19000$ of the images in the dataset for validation as shown in Table-{\ref{table23}}.
\begin{table}[t]
\centering

\caption{FLIR-ADAS and KAIST Multi-Spectral datasets partition topology for training and testing the proposed network.}
\label{table23}
\resizebox{9cm}{!}{%
\begin{tabular}{@{}l|l|c|c@{}}
\toprule
Dataset              & Total Images & \multicolumn{1}{l|}{Train Images} & \multicolumn{1}{l}{Test Images} \\ \midrule
FLIR-ADAS            & 10228        & 8862                              & 1366                            \\
KAIST Multi-Spectral & 95000        & 76000                             & 19000                           \\ \bottomrule
\end{tabular}%
}
\end{table}

\subsection{Object Detection in Thermal Images through Style Consistency (ODSC)}

The evaluation of the proposed method is demonstrated using state-of-the-art object detection networks. The object detection networks include Faster-RCNN, SSD-300, and SSD-512. These object detection networks are implemented with different backbone architecture; for instance, ResNet-101 is used as a backbone network in Faster-RCNN; VGG16, MobileNet, and EfficientNet are used with SSD-300; SSD-512 uses VGG16 as backbone architecture. 
The dataset comprises of FLIR-ADAS and KAIST Multi-Spectral dataset. The FLIR-ADAS dataset is partitioned into training and testing using a standard split, while the KAIST Multi-Spectral dataset is only used in testing the object detection networks.
All the networks are implemented in Pytorch, having formulated the data in PASCAL-VOC format. The standard PASCAL-VOC evaluation criteria are used in this study \cite{pascal}.

\subsubsection{Baseline}
A baseline approach is experimented first for the comparative analysis with the proposed methodology, which involves training and testing of object detection network using thermal images only. 
In training the Faster-RCNN, ResNet-101 backbone is  adapted and trained on the thermal image dataset. The network is trained using Adam optimizer with a learning rate of $10^{-4}$ and a momentum of $0.9$ for total of $15$ epochs. 

\par 
The experimental evaluation with the SSD object detection network constitutes two different architectures, i-e SSD-300 and SSD-512. In the case of training the SSD-300, the backbone networks are trained on the training data. The learning rate for VGG16, MobileNet, and EfficientNet used as the backbone network for SSD-300 are $10^{-4}$,$10^{-3}$, and $10^{-3}$, respectively. For the SSD-512 experimentation, only  VGG-16 is used as a backend for training with a learning rate of $10^{-3}$. All the networks have used a batch size of $4$ on the Nvidia-TITAN-X having $12$GB of computational memory.

\subsubsection{Experimental Configuration of ODSC}
In the proposed methodology, the MSGNet is trained with thermal images to serve as a content image, whereas the RGB images correspond to style images, as shown in Fig. \ref{style-images} (a). In training the MSGNet, VGG16 is used as a loss network. The pre-trained weights of the loss network on the ImageNet dataset are employed for training the MSGNet. In a loss network, the balancing weights as referred to in the Eq. \ref{equ3} are $\lambda_c=1$ and $\lambda_s=5$ respectively while the total variational regularization for content and style is $\lambda_{TV}=10^{-6}$. In the experimental configuration, the size of the style image $x_s$ is iteratively updated, having a size of  $256,512,768$, respectively. The size of the content images is resized to $256 \times 256$. The Adam optimizer is used with a learning rate of $10^{-3}$ in the training configuration. The MSGNet is trained for a total of $100$ epochs with a batch of $4$ on the Nvidia-TITAN-X.

\begin{table*}[b]
\centering
\caption{Quantitative analysis using Proposed Method (ODSC) configuration.}
\label{table-2}
\resizebox{12cm}{!}{%
\begin{tabular}{@{}llcccc|c@{}}
\toprule
\multicolumn{6}{c|}{FLIR ADAS Dataset} & KAIST Multi-Spectral Dataset \\ \midrule
Network Architecture & Backbone & car & bicycle & person & Average mAP & person \\ \midrule
Faster-RCNN & ResNet-101 & 0.7190 & 0.4394 & 0.6201 & \textbf{0.5928} & \textbf{0.5745} \\ \midrule
SSD-300 & VGG-16 & 0.7991 & 0.4691 & 0.6253 & \textbf{0.6312} & \textbf{0.7536} \\ \midrule
SSD-300 & MobileNet-v2 & 0.5434 & 0.2798 & 0.3638 & \textbf{0.3957} & \textbf{0.7465} \\ \midrule
SSD-300 & EfficientNet & 0.7405 & 0.3512 & 0.5169 & \textbf{0.5362} & \textbf{0.6770} \\ \midrule
SSD-512 & VGG-16 & 0.8233 & 0.5553 & 0.7101 & \textbf{0.6962} & \textbf{0.7725} \\ \bottomrule
\end{tabular}%
}
\end{table*}
\begin{table*}[t]
\centering
\caption{Quantitative analysis using Baseline configuration for object detection networks.}
\label{table-1}
\resizebox{12cm}{!}{%
\begin{tabular}{@{}llcccc|c@{}}
\toprule
\multicolumn{6}{c|}{FLIR ADAS Dataset} & KAIST Multi-Spectral Dataset \\ \midrule
Network Architecture & Backbone & car & bicycle & person & Average mAP & person \\ \midrule
Faster-RCNN & ResNet-101 & 0.6799 & 0.4276 & 0.548 & 0.5518 & 0.5583 \\ \midrule
SSD-300 & VGG-16 & 0.7561 & 0.4502 & 0.6197 & 0.6087 & 0.6687 \\ \midrule
SSD-300 & MobileNet-v2 & 0.4774 & 0.1943 & 0.3163 & 0.3284 & 0.5998 \\ \midrule
SSD-300 & EfficientNet & 0.6809 & 0.2747 & 0.4992 & 0.4849 & 0.6162 \\ \midrule
SSD-512 & VGG-16 & 0.8055 & 0.5399 & 0.702 & 0.6825 & 0.6409 \\ \bottomrule
\end{tabular}%
}
\end{table*}
\par 
The trained model of MSGNet results in the generation of style images, as shown in Fig. \ref{style-images} (a). These style images are used in training the object detection networks. The detection networks trained on style images are evaluated on the test data comprise of thermal images. The training configuration of these object detection networks is kept similar as the baseline configuration to make a comparative analysis. 


\subsubsection{Experimental Results}
\begin{figure*}[t]
      \centering
      \includegraphics[width=12cm]{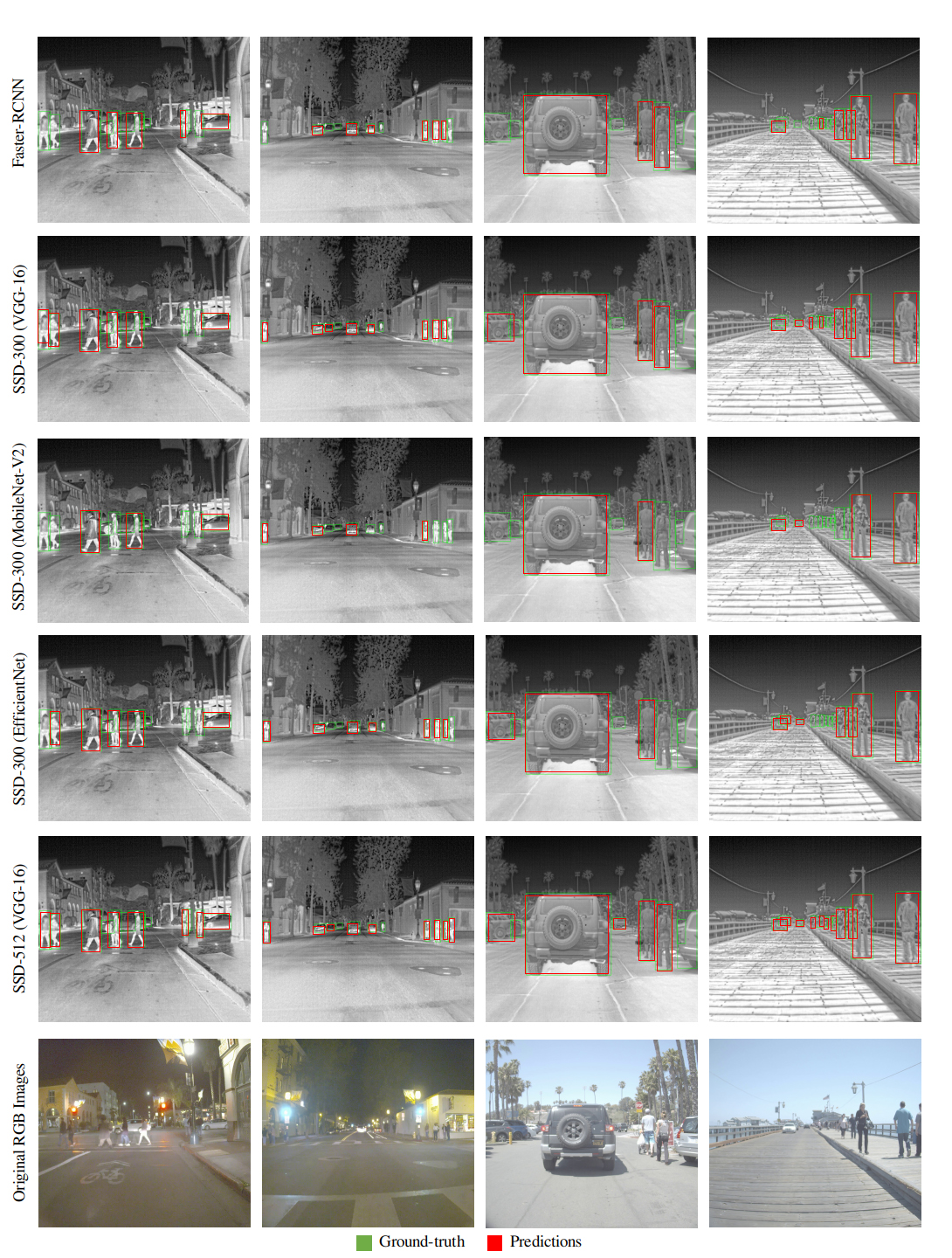}
      \caption{illustrates the qualitative results of object detection in thermal images through style consistency. The object detection results of all the detection networks are illustrated along with ground-truth and predictions on FLIR ADAS dataset.The second last row shows the The qualitative results of best model configuration (SSD512+VGG16). (Best viewed in color)}
      \label{OSDC-1}
\end{figure*}

\begin{figure*}[t]
      \centering
      \includegraphics[width=12cm]{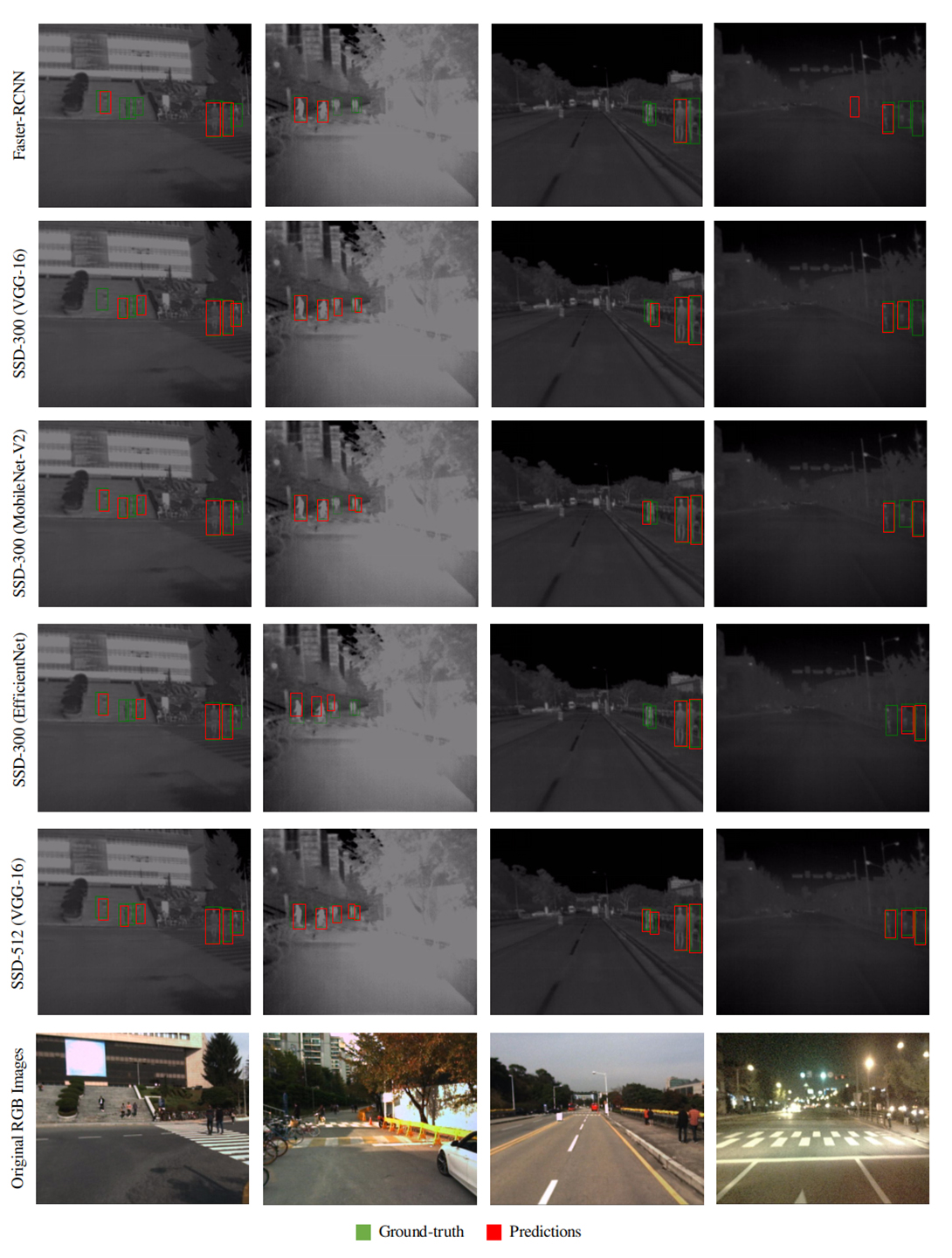}
      \caption{illustrates the qualitative results of object detection in thermal images through style consistency. The object detection results of all the detection networks are illustrated along with ground-truth and predictions on KAIST Multi-Spectral dataset.The second last row shows the The qualitative results of best model configuration (SSD512+VGG16). (Best viewed in color)}
      \label{OSDC-2}
\end{figure*}

For the evaluation of our experimental configuration, we have tested the baseline and proposed method, on both thermal datasets (FLIR ADAS and KAIST Multi-Spectral). Table-\ref{table-1} shows the mean average precision (mAP) scores of the baseline configuration for each detection network, i.e., the networks are trained on thermal images and evaluated on thermal images. Table-\ref{table-2} shows that the quantitative results of the proposed method. The best model configuration for the proposed method is (SSD512+VGG16) as shown in experimental results. The mAP score of the best model configuration of the proposed method has a better evaluation score compared to the baseline configuration. We perform a sanity check by conducting experiment by training network on thermal images and testing them on style images.  The detection networks trained on the thermal images tested on the style images show the marginal efficacy, as shown by Table-\ref{table-3}. Fig. \ref{OSDC-1}-\ref{OSDC-2} illustrate the qualitative result of object detection in thermal images through style consistency on FLIR ADAS and KAIST Multi-Spectral respectively for all the detection networks.
\begin{figure*}[t]
      \centering
      \includegraphics[width=12cm]{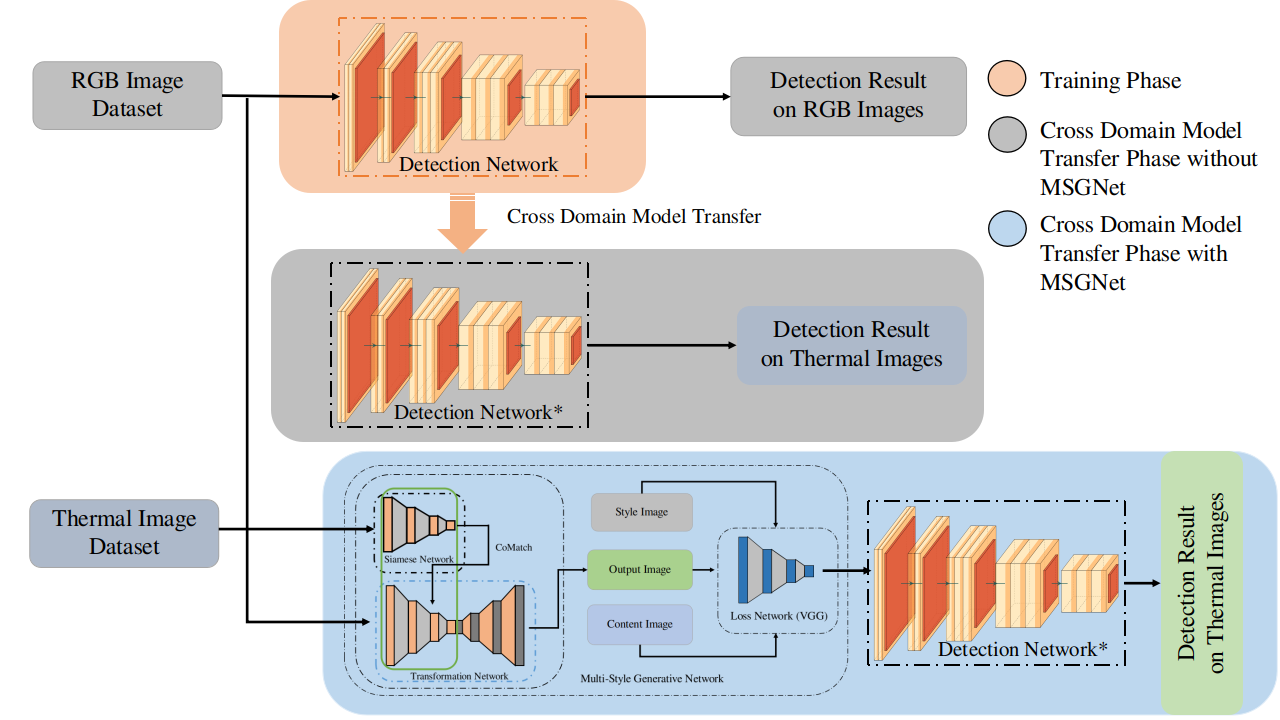}
      \caption{An overview of the cross-domain model transfer method. The detection networks are trained using the visible spectrum (RGB images). Afterward, these trained models are tested by implying the cross-model transfer with style transfer using MSGNet and also without style transfer. (Detection Network*) implies that the same detection networks are used for testing in the target domain.}
      \label{cross-domain}
\end{figure*}
\begin{table*}[b]
\centering
\caption{Quantitative analysis of testing object detection networks trained on thermal images and tested on style images}
\label{table-3}
\resizebox{12cm}{!}{%
\begin{tabular}{@{}llcccc|c@{}}
\toprule
\multicolumn{6}{c|}{FLIR ADAS Dataset} & KAIST Multi-Spectral Dataset \\ \midrule
Network Architecture & Backbone & car & bicycle & person & Average mAP & person \\ \midrule
Faster-RCNN & ResNet-101 & 0.3030 & 0.1985 & 0.2115 & 0.2377 & 0.1410 \\ \midrule
SSD-300 & VGG-16 & 0.6824 & 0.3286 & 0.5260 & 0.5123 & 0.6137 \\ \midrule
SSD-300 & MobileNet-v2 & 0.4551 & 0.1363 & 0.2899 & 0.2937 & 0.4773 \\ \midrule
SSD-300 & EfficientNet & 0.3637 & 0.1193 & 0.2289 & 0.2373 & 0.4449 \\ \midrule
SSD-512 & VGG-16 & 0.6779 & 0.3736 & 0.5538 & 0.5351 & 0.4961 \\ \bottomrule
\end{tabular}%
}
\end{table*}
\subsection{Corollary to proposed method: Cross Domain Model Transfer for Object detection in Thermal Images (CDMT)}
For the further investigation of the proposed method, a cross-domain model for thermal object detection is designed. The purpose of this study is to analyze the effect of trained RGB  detection models on styled and without styled images. It is to be noted that for cross-domain model transfer, the source and target domain are swapped compared to the first part of the proposed work. The reason of this configuration is to analyze the performance of object detectors that are trained on the RGB domain, when applied to thermal domain produce unsatisfactory results because of the fact of domain invariance.

However, if the style from the thermal domain is being employed on the content image of RGB domain, the trained RGB domain object detection networks performance improved since the style transfer bridge the gap between the two domains.  Fig.{\ref{cross-domain}} shows the overall framework for cross-domain model transfer object detection in thermal images. The detection networks ( Faster-RCNN backbone with ResNet-101, SSD-300 with backbone VGG16, MobileNet, and EfficientNet, SSD-512 with VGG16 backbone) are trained on the visible spectrum (RGB images) and then the trained model is tested on the thermal images.
As the detection networks are trained on a different domain, in this case, visible spectrum (RGB) images, the performance of these networks on thermal images will be marginal as can be seen in results.  The efficacy of thermal object detection can be increased by using the style consistency. The MSGNet is trained with RGB images as the content image, and the style is borrowed from the thermal images. The style transferred images are then passed to the same detection networks that are trained earlier on the visible spectrum (RGB) images, which improves the object detection in thermal style images. This cross-domain model transfer can be applied as a weak object detection module for the unlabeled dataset, as in our case for thermal images. 
\clearpage

\clearpage

\subsubsection{Experimental Configuration of CDMT}
The cross-domain model evaluation employs the training of object detectors on the visible spectrum (RGB images). The KAIST Multi-Spectral dataset is used in this experiment, considering that the labels are available for both domains. The object detection networks incorporated in this study include Faster-RCNN, SSD-300, and SSD-512. The network model configuration is similar to  ODSC. The Faster-RCNN is backend with ResNet-101 backbone. The SSD-300 network is experimented with VGG16, MobileNet, and EfficientNet backbone. Furthermore, SSD-512 is backend with VGG16 architecture. The learning rate for training all detection networks is $10^{-3}$ except for the SSD-300 with EfficientNet backbone, which is tested with $10^{-4}$. The batch size is $4$ for all the aforementioned detection networks.
\par
Similar to the ODSC, MSGNet is used to generate styled images, as shown by Fig.{\ref{style-images}}(b). In this case, the content images consist of the visible domain (RGB images), and the style is transferred from thermal images, which signifies that the style transfer between the content image (RGB images) and style image (thermal images) increase the object detection efficacy. The hyper-parameters for the MSGNet are kept the same as described in the experimental configuration of object detection in thermal images through style consistency. The detection networks are then tested on these generated styled images.
\subsubsection{Experimental Results}
The method's assessment is investigated by evaluating the trained network on the styled images and non-styled images (thermal images).
Table-\ref{table-4} shows the quantitative results of cross-domain model transfer. 
The quantitative results show that using the cross-domain model transfer with style transfer increases the object detection efficacy compared to cross-domain model transfer without style transfer. In addition to that, the method of using cross-domain model transfer will overcome the gap of annotating the unlabeled dataset and assists as a weak detector for the unlabeled dataset. The qualitative evaluation of using style transfer for CDMT is shown in Fig. \ref{CDMT-1} for all the detection networks.
\begin{table*}[t]
\centering
\caption{Quantitative analysis of Cross Domain Model Transfer (CDMT) }
\label{table-4}
\resizebox{12cm}{!}{%
\begin{tabular}{@{}llcc@{}}
\toprule
\multicolumn{4}{c}{KAIST Multi-Spectral Dataset}                                             \\ \midrule
                     & Domain       & CDMT without Style Transfer & CDMT with Style Transfer \\ \midrule
Network Architecture & Backbone     & person                      & person                   \\ \midrule
Faster-RCNN          & ResNet-101   & 0.5354                      & \textbf{0.7254}          \\ \midrule
SSD-300              & VGG-16       & 0.6098                    & \textbf{0.7598}          \\ \midrule
SSD-300              & MobileNet-v2 & 0.2512                      & \textbf{0.7012}          \\ \midrule
SSD-300              & EfficientNet & 0.1995                      & \textbf{0.5495}          \\ \midrule
SSD-512              & VGG-16       & 0.6202                      & \textbf{0.7702}          \\ \bottomrule
\end{tabular}%
}
\end{table*}

\begin{table*}[t]
\centering
\caption{Comparison of our proposed methods (ODSC and CDMT) with state-of-the-art methods.(*) represent average (day+night) mean Average Precision score. (-) indicates that the respective algorithm is not tested on the specified dataset.}
\label{table-5}
\resizebox{12cm}{!}{%
\begin{tabular}{@{}llcccc|c@{}}
\cmidrule(l){2-7}
\multicolumn{1}{l}{} & Dataset & \multicolumn{4}{c}{FLIR ADAS} & KAIST Multi-Spectral\\ \cmidrule(l){2-7} 
\multicolumn{1}{l}{} & Method & car & bicyle & person & mAP & person (mAP) \\ \cmidrule(l){2-7} 
\multicolumn{1}{l}{} & MMTOD-UNIT \cite{a10} & 0.7042 & 0.4581 & 0.5945 & 0.5856 & - \\ \cmidrule(l){2-7} 
\multicolumn{1}{l}{} & MMTOD-CG \cite{a10} & 0.6985 & 0.4396 & 0.5751 & 0.5711 & 0.5226 \\ \cmidrule(l){2-7} 
\multicolumn{1}{l}{} & PiCA-Net \cite{a9} & - & - & - & \textbf{-} & 0.658* \\ \cmidrule(l){2-7} 
\multicolumn{1}{l}{} & $R^3$Net \cite{a9} & - & - & - & \textbf{-} & 0.7085* \\ \cmidrule(l){2-7} 
\multicolumn{1}{l}{} & Intel \cite{a19} & 0.571 & 0.1312 & 0.245 & 0.3157 & - \\ \cmidrule(l){2-7} 
\multicolumn{1}{l}{} & tY model \cite{ty} & - & - & - & - & 0.630 \\ \cmidrule(l){2-7} 
\multicolumn{1}{l}{} & ACF+T+THOG \cite{KAIST} & - & - & - & - & 0.7139 \\ \midrule
\multirow{5}{*}{Ours (ODSC)} & Faster-RCNN+ResNet101 & 0.7190 & 0.4394 & 0.6201 & 0.5928 & 0.5345 \\ \cmidrule(l){2-7} 
 & SSD300 +VGG16 & 0.7991 & 0.4691 & 0.6253 & 0.6312 & 0.7536 \\ \cmidrule(l){2-7} 
 & SSD300+ Mobilenet V2 & 0.5434 & 0.2798 & 0.3638 & 0.3957 & 0.7465 \\ \cmidrule(l){2-7} 
 & SSD300+ EfficientNet & 0.7405 & 0.3512 & 0.5169 & 0.5362 & 0.6770 \\ \cmidrule(l){2-7} 
 & SSD512+VGG16 & \textbf{0.8233} & \textbf{0.5553} & \textbf{0.7101} & \textbf{0.6962} & \textbf{0.7725} \\ \midrule
\multirow{5}{*}{Ours (CDMT)} & Faster-RCNN+ResNet101 & - & - & - & - & 0.7254 \\ \cmidrule(l){2-7} 
 & SSD300 +VGG16 & - & - & - & - & 0.7598 \\ \cmidrule(l){2-7} 
 & SSD300+ Mobilenet V2 & - & - & - & - & 0.7012 \\ \cmidrule(l){2-7} 
 & SSD300+ EfficientNet & - & - & - & - & 0.5495 \\ \cmidrule(l){2-7} 
 & SSD512+VGG16 & - & - & - & - & \textbf{0.7702} \\ \bottomrule
\end{tabular}%
}
\end{table*}

\begin{figure}[t]
      \centering
      \includegraphics[width=9cm]{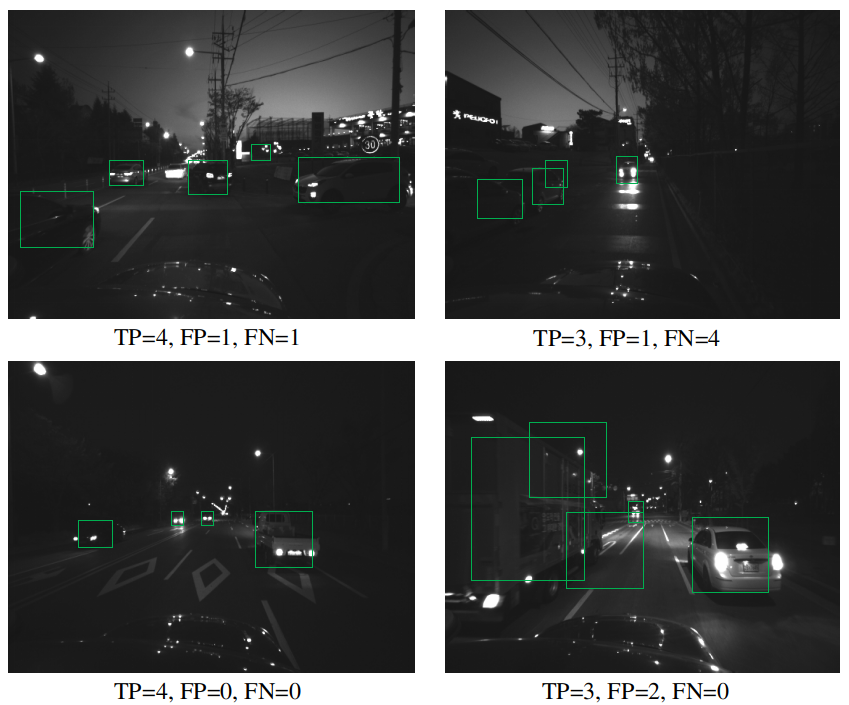}
      \caption{The illustration of weak label annotation using the cross-domain model transfer performed on our collected unlabeled dataset.}
      \label{cd-acc}
\end{figure}
\section{Discussion}
For the efficacy of the proposed methodology, an extensive analysis is conducted using state-of-the-art methods. \mbox{\cite{KAIST}} proposed multispectral aggregated channel features to detect the pedestrian in thermal images , which are tested with a limited domain of objects, whereas the objective of this study is to extend object detection for self-driving vehicles.  A detailed comparison of the state-of-the-art deep neural networks used for person detection in thermal images is provided in \mbox{\cite{ty}}, and a detailed comparison of Faster-RCNN with three different thermal images dataset is provided in \mbox{\cite{a19}} which is evaluated specifically for self-driving vehicles. \mbox{\cite{a9}} augments multispectral images with their saliency map to employ an attention mechanism that focuses attention on pedestrians during the daytime. They have trained the Faster-RCNN for pedestrians detection and fine-tuned it on extracted feature maps. \mbox{\cite{a10}} has used CycleGAN to generate the thermal images from RGB images, to remove the dependency of pairing 

\afterpage{
\begin{figure*}[t]
      \centering
      \includegraphics[width=12cm]{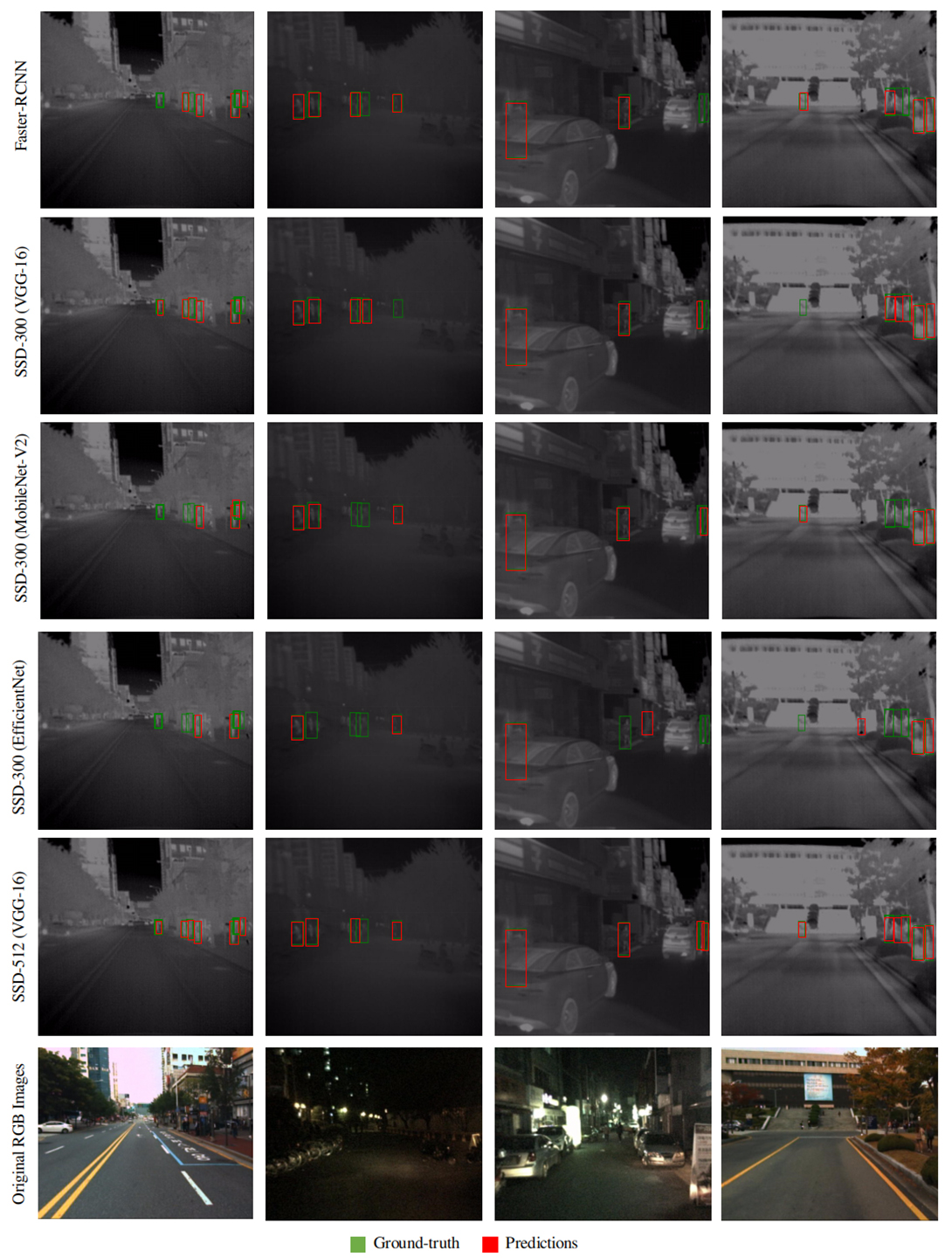}
      \caption{Object detection results using cross domain model transfer is illustrated. The ground-truth and predictions results of all the detection network are shown. The second last row shows the The qualitative results of best model configuration (SSD512+VGG16). (Best viewed in color) }
      \label{CDMT-1}
\end{figure*}
\clearpage
}
\noindent the RGB and thermal images in the dataset. They have used a variant of Faster-RCNN, which used both the thermal and RGB images to detect objects.
However, in comparison to these methods, we have adopted a novel framework to transfer low-level features from the source domain to the target domain at the image data level and have trained a deep neural network for thermal object detection.  
Table-\ref{table-5} shows a comparison between the proposed methods (ODSC and CDMT) and state-of-the-art methods. Our analysis has considered those methods in which the standard PASCAL-VOC evaluation is used for both FLIR-ADAS and KAIST Multi-Spectral datasets. 
\par 
In addition to the mAP scores, class mAP scores are also compared with state-of-the-art methods compared to the proposed approach. Further, the proposed method's comparison is not limited to the methods that only include domain adaptation. The object detection results are compared with the general object detection methods like PiCA-Net \cite{a9} and R$^3$Net \cite{a9}, which have used saliency maps for object detection for thermal images. It is apparent from the Table-\ref{table-5} illustrates that in most of the categories, our proposed strategies have better performance efficacy as compared to the existing benchmark.

\par
The inference frame rate for the detection neural network used in the proposed method is illustrated in Table-{\ref{fps}}. The number of frames per second is calculated on the Nvidia-TITAN-X having $12$GB of memory. For the cross-domain model transfer usage in the context of weak labeler for the unlabeled dataset, we have experimented with our own unlabeled dataset collected using i3 systems TE-EQ1 / TE-EV1 \footnote{http://i3system.com/uncooled-detector/te-eq1/?lang=en} thermal camera. Fig.{\ref{cd-acc}} illustrates the weak label annotation performed by cross-domain model transfer. The results only illustrate the true positive (TP), false positive (FP) and false-negative (FN). The overall accuracy on the whole unlabeled dataset is $67.36\%$.

\begin{table}[t]
\centering
\caption{The inference frame per second evaluation of deep neural networks models used in the proposed work.}
\label{fps}
\resizebox{7cm}{!}{%
\begin{tabular}{@{}l|c@{}}
\toprule
Network Architecture   & \multicolumn{1}{l}{Frame per second} \\ \midrule
Faster-RCNN -ResNet101 backbone & 9                                    \\
SSD300 -VGG16 backbone          & 31                                   \\
SSD300 -MobileNetv2 backbone    & 23                                   \\
SSD300 -EffecientNet backbone    & 16                                   \\
SSD512 -VGG16 backbone          & 11                                   \\ \bottomrule
\end{tabular}%
}
\end{table}


\section{Conclusion}
This study proposes a domain adaptation framework for object detection in underexposure regions for autonomous driving. The framework uses domain adaptation from visible domain to thermal domain through style consistency and utilizes MSGNet to transfer low-level features from the source domain to the target domain which keeps high-level semantic features intact. The proposed method outperforms the existing benchmark for object detection in the thermal images. Moreover, the effectiveness of style transfer is strengthened by using a cross-domain model transfer between visible and thermal domains. 
\par
The application of the proposed framework is found in the autonomous driving under low lighting conditions. Object detection is an integral to the core of perception and failure to detect an object compromises the safety of the autonomous driving. Thermal images provide additional meaningful data from the surroundings, and the proposed framework improves the results of object detection in thermal images consequently improving the safety of autonomous driving. In future work, we aim to integrate lane detection and segmentation into the proposed framework using thermal images. 

\section*{Acknowledgments}
This work was partly supported by Institute of Information \& Communications Technology Planning \& Evaluation (IITP) grant funded by the Korea government (MSIT) (No.2014-3-00077, AI National Strategy Project) and the National Research Foundation of Korea (NRF) grant funded by the Korea government (MSIT) (No. 2019R1A2C2087489), and Ministry of Culture, Sports and Tourism (MCST), and Korea Creative Content Agency (KOCCA) in the Culture Technology (CT) Research \& Development (R2020070004) Program 2020.

\end{document}